\documentclass[lettersize,journal]{IEEEtran}
\usepackage{amsmath,amsfonts,amssymb,booktabs}
\usepackage{algorithmic}
\usepackage{algorithm}
\usepackage{array}
\usepackage[caption=false,font=normalsize,labelfont=sf,textfont=sf]{subfig}
\usepackage{textcomp}
\usepackage{stfloats}
\usepackage[hidelinks]{hyperref}
\usepackage{url}
\usepackage{verbatim}
\usepackage{graphicx}
\usepackage{cite}
\hypersetup{
  colorlinks   = true, 
  urlcolor     = blue, 
  linkcolor    = blue, 
  citecolor   = red 
}

\begin{document}

\title{Graph Neural Networks for Contextual ASR with the Tree-Constrained Pointer Generator}

\author{Guangzhi Sun,~\IEEEmembership{Student Member,~IEEE,} Chao Zhang,~\IEEEmembership{Member,~IEEE,} Philip C. Woodland,~\IEEEmembership{Fellow,~IEEE}
\IEEEcompsocitemizethanks{
\IEEEcompsocthanksitem G. Sun and P.C. Woodland are with the Department of Engineering, University of Cambridge, Trumpington St., Cambridge.
Email: \{gs534, pcw\}@eng.cam.ac.uk
\IEEEcompsocthanksitem C. Zhang is with the Department of Electronic Engineering, Tsinghua University, Beijing, China.
Email: cz277@tsinghua.edu.cn
}
}


 
\maketitle

\begin{abstract}
The incorporation of biasing words obtained through contextual knowledge is of paramount importance in automatic speech recognition (ASR) applications. This paper proposes an innovative method for achieving end-to-end contextual ASR using graph neural network (GNN) encodings based on the tree-constrained pointer generator method. GNN node encodings facilitate lookahead for future word pieces in the process of ASR decoding at each tree node by incorporating information about all word pieces on the tree branches rooted from it. This results in a more precise prediction of the generation probability of the biasing words. The study explores three GNN encoding techniques, namely tree recursive neural networks, graph convolutional network (GCN), and GraphSAGE, along with different combinations of the complementary GCN and GraphSAGE structures. The performance of the systems was evaluated using the Librispeech and AMI corpus, following the visual-grounded contextual ASR pipeline. The findings indicate that using GNN encodings achieved consistent and significant reductions in word error rate (WER), particularly for words that are rare or have not been seen during the training process. Notably, the most effective combination of GNN encodings obtained more than 60\% WER reduction for rare and unseen words compared to standard end-to-end systems.
\end{abstract}


\section{Introduction}
\IEEEPARstart{E}{nd-to-end} ASR systems often struggle with the accurate recognition of rare ``long-tail" words that were not present in the training data. To combat this issue, Contextual biasing which applies external contextual knowledge to the ASR system during inference, becomes a crucial factor in addressing the long-tail word problem in various applications \cite{shallow_context_1,shallow_context_2,shallow_context_3,deep_context_1,deep_context_2,deep_context_3,deep_context_4,deep_context_5,deepshallow,DBRNNT,ne_correction,unsupervised_context,word_mapping,lm_pointer}. Contextual knowledge is often represented as a list (referred to as a \textit{biasing list}) of words or phrases (referred to as \textit{biasing words}) that are likely to appear in a given context. Biasing lists can be sourced from various resources, such as a user's contact book or playlist, recently visited websites and visual information from presentation slides \textit{etc}. 
Despite their infrequent occurrence and thus the limited influence on the overall word error rate (WER), biasing words significantly impacts understanding the content as biasing words are mostly content words such as nouns or proper nouns, which are crucial for downstream tasks. The inclusion of a word in a biasing list increases its likelihood of being correctly recognised, making contextual biasing a crucial component for the accurate recognition of rare content words.

End-to-end trainable ASR systems \cite{e2e_attention_1, e2e_rnnt_1} are designed to encapsulate all necessary knowledge within a single, static model, making it difficult to integrate dynamic context-specific knowledge at test-time. To overcome this challenge, specialised contextual biasing methods have been proposed, such as shallow fusion (SF) with a weighted finite-state transducer (WFST) or an adapted language model (LM) that incorporates contextual knowledge \cite{shallow_context_1,shallow_context_2,shallow_context_3,lm_pointer,unsupervised_context,word_mapping}, attention-based deep context approaches \cite{deep_context_1,deep_context_2,deep_context_3,deep_context_4,deep_context_5}, as well as deep biasing (DB) with a prefix tree for improved efficiency when dealing with large biasing lists \cite{deep_context_3,deepshallow,DBRNNT}. More recently, contextual biasing components with a pointer generator mechanism \cite{pointer_1,pointer_2,pointer_3} that directly modifies the output distribution have been proposed \cite{TCPGen, MEM}, which can be jointly optimised with ASR systems. In particular, the tree-constrained pointer generator (TCPGen) component proposed in \cite{TCPGen} builds a neural shortcut by directly interpolating the original model distribution with the TCPGen distribution estimated from contextual knowledge structured as a prefix-tree, based on a dynamic interpolation weight predicted by the TCPGen component. TCPGen performance was further boosted by encoding the prefix-tree using a tree recursive neural network (tree-RNN) \cite{tcpgengnn}.

This paper substantially extends the work in \cite{tcpgengnn} and proposes to use three types of graph neural networks (GNN) for prefix-tree encoding in TCPGen \footnote{The main code is at \url{https://github.com/BriansIDP/espnet/tree/GNN}.}. These include tree-RNN, graph convolutional network (GCN) \cite{gcn} with its variant GCNII \cite{gcnii}, and GraphSAGE with the max-pooling aggregator \cite{sage}. While tree-RNN is a representative GNN model with a single recursive layer, GCN and GraphSAGE are two popular and effective multi-layer GNN designs. Specifically, GCN encodes the tree by utilising spectral representation, while GraphSAGE, as a spatial method, directly explores the graph topology \cite{gnnreview}. To further enhance the performance of GNN tree encodings, this paper proposes attentive and bilinear combination approaches to exploit the complementarity between GCN and GraphSAGE. Additionally, this paper introduces an effective parameter-tying scheme for both GCN and GraphSAGE to improve their performance with deeper structures.

GNN encodings provide more powerful node representations in the prefix tree of TCPGen, allowing for ``lookahead" functionality where each node contains not only its own word piece information but also information about its child branches. This improved node representation in TCPGen leads to more accurate generation probability predictions for biasing words, enabling better contextual biasing by incorporating information about future word pieces during each ASR decoding step. TCPGen with GNN encodings, as a generic component for end-to-end ASR, is integrated into both attention-based encoder-decoder (AED) \cite{e2e_attention_1,e2e_attention_2,e2e_attention_3,e2e_attention_4,e2e_attention_5,LAS} and neural transducer architecture (N-T) \cite{e2e_rnnt_1,e2e_rnnt_2,e2e_rnnt_3,e2e_rnnt_4,e2e_rnnt_5}. 

Experiments were conducted with two different setups: (1). a simulated contextual ASR task using LibriSpeech audiobook data, and,  (2). an audio-visual speech recognition pipeline \cite{tcpgengnn} with the AMI meeting data. In addition to the consistent and large reductions in error rates achieved by TCPGen with tree-RNN in \cite{tcpgengnn}, using the proposed GNN encodings, especially combined ones achieved further significant improvements in the word error rate (WER) of rare content words. 

The remainder of this paper is organised as follows. Section \ref{sec:related} reviews related work. Section \ref{sec:tcpgen} introduces the TCPGen component. Section \ref{sec:gnn} describes the details of applying proposed GNNs. Section \ref{sec:exp} and \ref{sec:results} present the experimental setup and results. Section \ref{sec:conclusion} gives conclusions.

\section{Related work}
\label{sec:related}
\subsection{End-to-end contextual speech recognition}
Recently, various contextual biasing algorithms have been developed for end-to-end ASR. One prominent research stream focuses on representing biasing lists as external weighted finite-state transducers (WFSTs), which are integrated into a class-based language model (LM) via shallow fusion (SF) \cite{shallow_context_1,shallow_context_2,shallow_context_3,word_mapping}. These methods often depend on context prefixes such as ``call" or ``play", limiting their ability to handle the diverse grammar in natural speech. On the other hand, deep context approaches, often using attention mechanisms, have also been proposed, which encode the biasing list into a vector to use as input for the end-to-end ASR models \cite{deep_context_1,deep_context_2,deep_context_3,deep_context_4,deep_context_5}. While deep context approaches eliminate the dependence on syntactic prefixes seen in SF methods, they require more memory and are less effective for handling large biasing lists.

The study in \cite{deepshallow} combined the use of deep context and shallow fusion of a WFST in an N-T, leading to improved efficiency by limiting the biasing vector extraction to a subset of word pieces determined by a prefix tree representation of the biasing list, which is referred to as deep biasing (DB). The prefix-tree-based method was further expanded in \cite{DBRNNT} to include RNN LMs for shallow fusion, resulting in improved recognition of biasing words. Prior research only analysed industry datasets, however, \cite{DBRNNT} proposed and validated a simulation of contextual biasing on open-source data by incorporating a large number of distractors into the list of biasing words in the utterance. More recently, \cite{MEM} and \cite{TCPGen} simultaneously proposed a neural shortcut between the biasing list and the final model output distribution. In contrast to \cite{MEM}, TCPGen \cite{TCPGen} adopted a structured prefix-tree representation for biasing lists, which also enabled the future development of tree-RNN encodings \cite{tcpgengnn}.

\subsection{GNN for speech and language processing}
GNNs have been extensively employed in a multitude of speech and language tasks. In language-related applications, such as sentence-level text classification or word-level sequence labelling, GNNs are often utilized to capture the syntactic dependencies or semantic relations among words in a sentence. Furthermore, the encoding of subword-unit-based tree structures using GNNs \cite{RNNsubword,RNNsubword2} has been explored for the purpose of generating more effective word representations. GNNs have also been applied to named-entity recognition \cite{RNNNER} and neural machine translation  \cite{RNNtranslation1, RNNtranslation2}.

GNNs also have numerous applications in speech processing, such as text-to-speech synthesis, where GNNs model the syntactic and semantic relationships in the text. In GraphTTS \cite{graphtts}, the authors structured the sentence into a hierarchical tree by dividing the utterance into words and then further into characters. This allowed the system to capture prosodic relationships among different parts of the input. Additionally, GNNs are used in paralinguistic tasks as the syntactic encoder, including sentiment classification and hate speech detection tasks \cite{RNNsemantic1, RNNsemantic2, RNNsentiment1}. In \cite{tcpgengnn}, a tree-RNN structure was used to encode a word piece prefix-tree in the TCPGen component for contextual biasing.

\section{Tree-constrained pointer generator}
\label{sec:tcpgen}

\begin{figure}[h]
    \centering
    \includegraphics[scale=0.25]{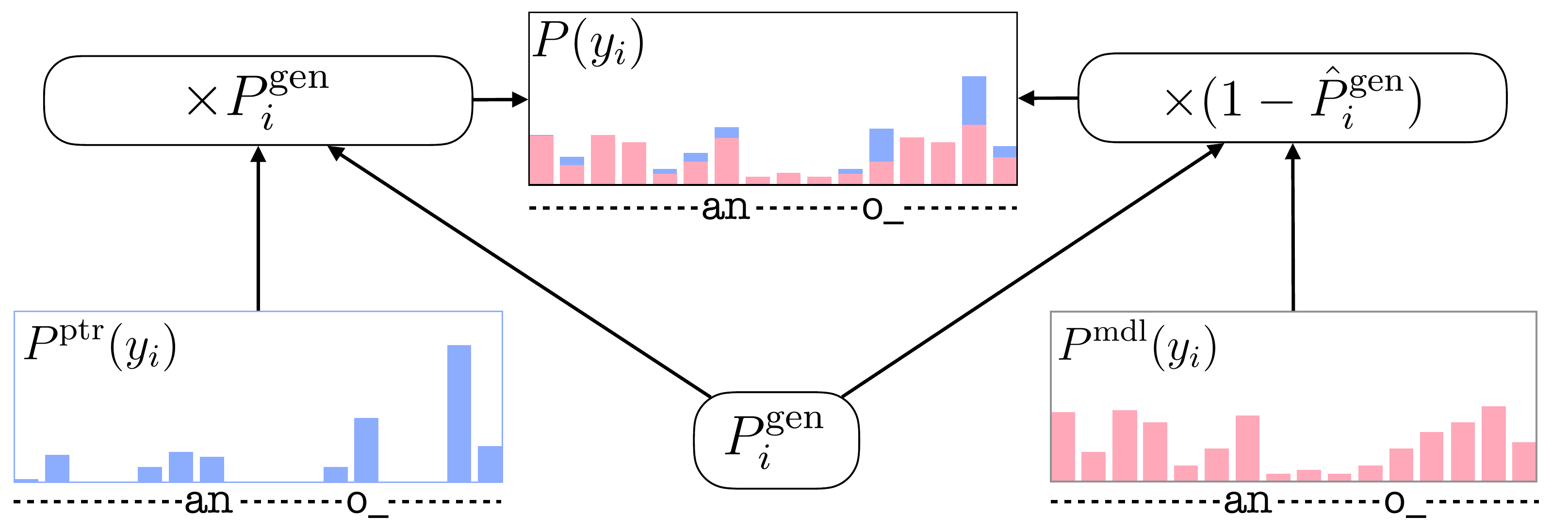}
    \caption{Illustration of interpolation in TCPGen with corresponding terms in Eqn. (\ref{eq:TCPGen_final}). $P^\text{ptr}(y_i)$ is the TCPGen distribution. $P^\text{mdl}(y_i)$ is the distribution generated by a standard end-to-end model. $P(y_i)$ is the final output distribution. $\hat{P}^\text{gen}_i$ and $P^\text{gen}_i$ are the scaled and unscaled generation probabilities.}
    \label{fig:interpolation}
\end{figure}

TCPGen is a neural network-based module that employs a combination of symbolic prefix-tree search and a neural pointer generator for contextual biasing, allowing for end-to-end optimisation. The biasing list is structured as a prefix tree at the word piece-level. At each output stage, TCPGen computes a probability distribution over a set of valid word pieces that are constrained by the prefix tree. TCPGen also predicts a dynamic generation probability, signifying the amount of contextual biasing required at the specific output step. The final output distribution is obtained by taking a weighted summation of the TCPGen distribution and the original AED or N-T output distribution (see Figure \ref{fig:interpolation}).

\begin{figure}[h]
    \centering
    \includegraphics[scale=0.25]{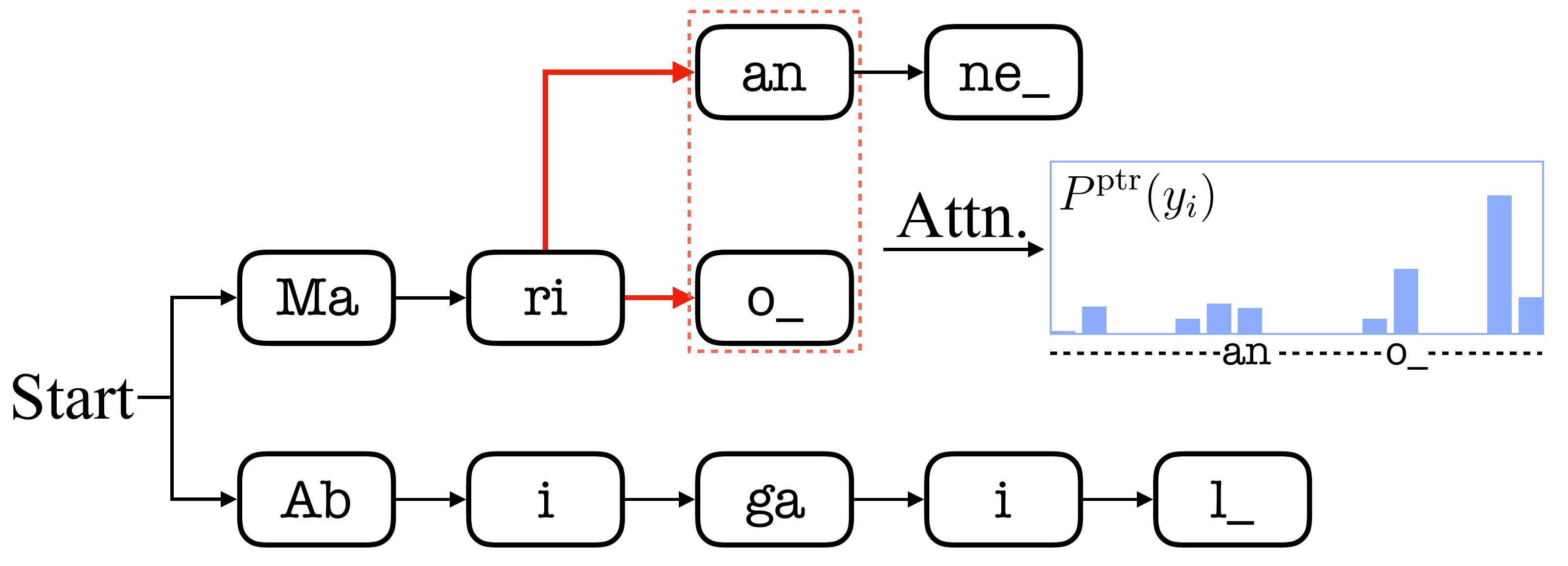}
    \caption{An example of prefix tree search and attention in TCPGen. With previous output \texttt{Tur}, \texttt{in\_} and \texttt{n} are two valid word pieces on which attention will be performed. A word end unit is denoted by \texttt{\_}.}
    \label{fig:tree}
\end{figure}

The key symbolic representation of the external contextual knowledge in TCPGen is the prefix tree. For simplicity, examples and equations in this section are presented for a specific search path, which can be generalised easily to beam-search with multiple paths. In the example prefix tree with biasing words (\texttt{turner}, \texttt{vignette} and \texttt{turin}) shown in Fig. \ref{fig:tree}, if the previously decoded word piece is \texttt{Tur}, word pieces \texttt{in\_} and \texttt{n} form the set of valid word pieces $\mathcal{Y}^\text{tree}_i$. 
Denoting $\mathbf{x}_{1:T}$ and $y_i$ as input acoustic features and output word piece, $\mathbf{q}_i$ as the query vector carrying the decoding history and acoustic information, $\mathbf{K}=[...,\mathbf{k}_j,...]$ as the key vectors, scaled dot-product attention is performed between $\mathbf{q}_i$ and $\mathbf{K}$ to compute the TCPGen distribution $P^\text{ptr}$ and the output vector $\mathbf{h}^{\text{ptr}}_i$ as shown in Eqns. \eqref{eq:TCPGen_attention} and \eqref{eq:TCPGen_value}.
\vspace{-0.1cm}
\begin{equation}
    P^{\text{ptr}}(y_{i}|y_{1:i-1},\mathbf{x}_{1:T}) = \text{Softmax}(\text{Mask}(\mathbf{q}_i\mathbf{K}^\text{T}/\sqrt{d})),
    \label{eq:TCPGen_attention}
    \vspace{-0.1cm}
\end{equation}
\begin{equation}
    \mathbf{h}^{\text{ptr}}_i = \sum\nolimits_{j} P^{\text{ptr}}(y_i=j|y_{1:i-1},\mathbf{x}_{1:T})\,\mathbf{v}^\text{T}_j,
    \label{eq:TCPGen_value}
    \vspace{-0.1cm}
\end{equation}
where $d$ is the size of $\mathbf{q}_i$ (see \cite{transformer}), Mask$(\cdot)$ sets the probabilities of word pieces that are not in $\mathcal{Y}^{\text{tree}}_i$ to zero, and $\mathbf{v}_j$ is the value vector relevant to $j$. 

TCPGen can be employed in both AED and N-T. In AED, the query is the combination of the context vector and the embedding of the preceding word piece, while the keys and values are derived from the decoder word piece embedding using a shared projection matrix. The generation probability in AED is computed from the concatenation of decoder hidden states and TCPGen output vectors $\mathbf{h}^{\text{ptr}}_i$, followed by Sigmoid activation function to be constrained to $(0, 1)$. In N-T, the pointer generator is applied to each combination of the encoder and the predictor step, where the TCPGen distribution is calculated using the concatenation of the corresponding encoder and predictor hidden states as the query. Keys and values in N-T are computed from the predictor word piece embeddings. The generation probability for N-T is derived from the joint network output and the TCPGen output vector $\mathbf{h}^{\text{ptr}}_i$. To ensure that the probability of the null symbol in N-T is unchanged, $P^{\text{ptr}}_i(\varnothing|\mathbf{x}_{1:T}, y_{1:i-1})$ is set to 0 and the generation probability is scaled by $1-P^{\text{mdl}}_i(\varnothing|\mathbf{x}_{1:T}, y_{1:i-1})$ where $P^{\text{mdl}}_i$ is the original model distribution before interpolation. 

For better flexibility, an \textit{out-of-list} (OOL) token is included in $\mathcal{Y}^{\text{tree}}_i$ indicating that no suitable word piece can be found in the set of valid word pieces. To ensure that the final distribution sums to 1, the generation probability, $P^{\text{gen}}_i$, is scaled as $\hat{P}^\text{gen}_i=P^\text{gen}_i(1-P^\text{ptr}(\text{OOL}))$, and the final output can be calculated as shown in Eqn. \eqref{eq:TCPGen_final}.
\begin{equation}
    P(y_i) = P^{\text{mdl}}(y_i)(1-\hat{P}^\text{gen}_i) + P^{\text{ptr}}(y_i)P^\text{gen}_i,
    \label{eq:TCPGen_final}
\end{equation}
where dependencies, $y_{1:i-1}, \mathbf{x}_{1:T}$, are omitted for clarity. $P^{\text{mdl}}(y_i)$ represents the output distribution from the standard end-to-end model.

\subsection{Biasing-driven LM discounting (BLMD) for TCPGen}
Log-linear interpolation is often used as a technique to incorporate an external LM via SF. Define the source domain data as the text of the training data for the end-to-end model, and the target domain data as the data used to train an external LM such that it generates better probability estimates for the test data. The LM discounting is defined as Eqn. \eqref{eq:lmdisc}.
\begin{align}
    P^{\text{sf}}(y_i) & = P^\text{mdl}(y_i)\frac{P^\text{tgt}(y_i)^{\alpha}}{P^\text{src}(y_i)^{\beta}},
    \label{eq:lmdisc}
\end{align}
where $P^\text{mdl}(Y)$ is the probability from the end-to-end system, $P^\text{src}(Y)$ is the probability of the source domain LM and $P^\text{tgt}(Y)$ is the target domain LM probability. Extending this idea to contextual biasing with TCPGen, BLMD can be applied as shown in Eqn. \eqref{eq:lmdisc2}.
\begin{align}
    P^{\text{sf}}(y_i) & = (1-P^\text{gen})P^\text{mdl}(y_i)\frac{P^\text{tgt}(y_i)^{\alpha_1}}{P^\text{src}(y_i)^{\beta_1}} \nonumber\\ & + P^\text{gen}P^\text{ptr}(y_i)\frac{P^\text{tgt}(y_i)^{\alpha_2}}{P^\text{src}(y_i)^{\beta_2}},
    \label{eq:lmdisc2}
\end{align}
where $P^\text{ptr}$ is the TCPGen distribution, and the same source and target LMs are used for both distributions, but with different sets of hyper-parameters $\alpha_1, \beta_1$ and $\alpha_2, \beta_2$ tuned on the validation set.

\section{GNN Tree Encodings for TCPGen}
\label{sec:gnn}
\begin{figure*}[t]
    \centering
    \includegraphics[scale=0.27]{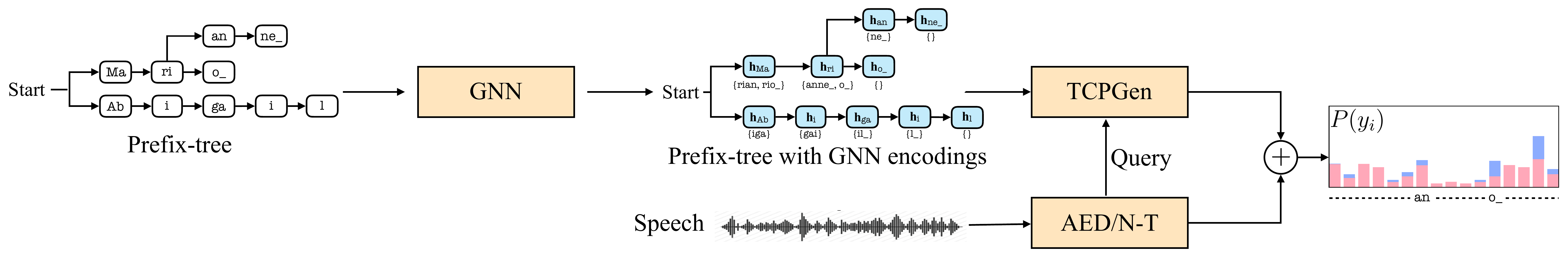}
    \caption{Pipeline of encoding prefix-tree with GNN for TCPGen. The prefix-tree is first encoded by a GNN, and the GNN-encoded tree is used by TCPGen to generate the TCPGen distribution where key and value vectors are GNN-based node encodings. The lookahead content for a 2-layer GCN as an example is denoted in \texttt{\{\}}, i.e. for node \texttt{ri}, $\mathbf{h}_\texttt{ri}$ covers information about \texttt{an ne\_} and \texttt{o\_}}
    \vspace{-0.3cm}
    \label{fig:pipeline}
\end{figure*}

While looking ahead into future branches of the paths being searched on the prefix tree is greatly beneficial to the correct prediction of the generation probability, node representations in standard TCPGen only contain information about the word piece on that node. To achieve lookahead functionality, GNNs are used to encapsulate future branch information into each node representation. The pipeline of applying GNN encodings in TCPGen is shown in Fig. \ref{fig:pipeline}.

The word piece prefix-tree is first encoded with a GNN to obtain encodings associated with each node. Then, the tree with GNN encodings is used by TCPGen, where the key and value for the TCPGen distribution are computed using the encoding of nodes in the set of valid word pieces, in place of word piece embeddings as shown in Eqn. \eqref{eq:TCPGeninlas_key}.
\vspace{-0.1cm}
\begin{equation}
    \mathbf{k}_j = {W}^{\text{K}}\mathbf{h}^\text{gnn}_{n_j} \hspace{1cm} \mathbf{v}_j = {W}^{\text{V}}\mathbf{h}^\text{gnn}_{n_j},
    \label{eq:TCPGeninlas_key}
    \vspace{-0.1cm}
\end{equation}
where ${W}^{\text{V}}$ and ${W}^{\text{K}}$ are parameter matrices, and $\mathbf{h}^\text{gnn}$ is the GNN node encoding obtained using different types of GNN. This paper explores three different types of GNNs, namely the tree-RNN, GCN (including its variant, GCNII) and GraphSAGE with max pooling, together with combinations of GCN and GraphSAGE as two complementary types of GNN. Details of GNN structures applied in TCPGen together with modifications are described in the following sections.

\subsection{Tree-RNN}
Tree-RNN recursively encodes the tree from leaf nodes to the root using an RNN structure. Specifically, at node $n_j$ which contains child nodes $n_1,...,n_k,...,n_K$, the vector representation of $n_j$ can be written as Eqn. \eqref{eq:treenet}.
\vspace{-0.1cm}
\begin{equation}
    \mathbf{h}^\text{trnn}_{n_j} = \text{ReLU}({W}_1\mathbf{y}_j + \sum_{k=1:K}{W}_2\mathbf{h}^\text{trnn}_{n_k}),
    \label{eq:treenet}
\end{equation}
where $\mathbf{h}^\text{trnn}_{n_k}$ is the vector representation of node $n_k$, and $\mathbf{y}_j$ is the embedding vector of the word piece of node $n_j$. ${W}_1$ and ${W}_2$ are parameter matrices jointly optimised with the ASR system by allowing gradient back-propagation through $\mathbf{h}^\text{trnn}_{n_k}$. In this way, each node recursively encodes information from its child nodes, such that the information of the entire branch rooted from it can be incorporated in the node encoding $\mathbf{h}^\text{trnn}_{n_k}$. 

Before the forward pass of the main ASR model, the encoding of each node is obtained by applying Eqn. \eqref{eq:treenet} recursively from leaf to root. Then, for the same example shown in Fig. \ref{fig:tree}, at the node of \texttt{Tur}, node encodings $\mathbf{h}^\text{trnn}_\texttt{in\_}$ and $\mathbf{h}^\text{trnn}_\texttt{n}$ are used to calculate the TCPGen distribution and $\mathbf{h}^\text{ptr}_i$. Therefore, if \texttt{Turner} appears in the utterance, TCPGen is aware of this entire word as early as in the encoding of \texttt{Tur}. Such lookahead functionality achieves a more accurate prediction of the generation probability to determine when contextual biasing is needed.

Although Tree-RNN achieves the lookahead functionality, it uses a rather simple RNN structure to encode the information of all succeeding nodes on that branch into a single vector representation. Therefore, more powerful and flexible GNN encodings are explored in order to improve performance.

\subsection{Graph Convolutional Network (GCN)}

As an alternative method to Tree-RNN, GCN is applied for tree encodings to achieve better node representations with controllable lookahead distance. GCN is a multi-layer network where each layer computes the encoding of a node as a function of its neighbours based on the graph Laplacian matrix. Each layer of GCN conducts one message passing from immediate neighbouring nodes. For a GCN with $L$ layers, the encoding of a node covers information from a node that is an $L$-hop ahead on branches rooted from it. 

Specifically, define $H^\text{gcn}(l) = [\mathbf{h}^\text{gcn}_{n_1}(l), ..., \mathbf{h}^\text{gcn}_{n_N}(l)]$ as the node encoding matrix of layer $l$ whose rows $\mathbf{h}^\text{gcn}_{n_j}(l)$ are the encoding of the $n_j$ nodes, the GCN layer computation is 
\begin{equation}
    H^\text{gcn}(l+1) = f(\hat{P}H^\text{gcn}(l)W{(l)}),
    \label{eq:gcn2}
\end{equation}
where $W(l)$ is the parameter matrix of $l$, $f(\cdot)$ is the activation function, $\hat{A}=A+I_N$ is the adjacency matrix with self-loops to enable the information of the current node to be included in the node representation, and $\hat{D}$ is the degree matrix of $\hat{A}$. A specific form of normalised graph Laplacian \cite{gcn}, 
\begin{equation}
    \hat{P}=\hat{D}^{-1/2}\hat{A}\hat{D}^{-1/2},
    \label{eq:gcn2a}
\end{equation}
is used to address the vanishing/exploding gradient problem. Note that as only future branch information is needed in TCPGen, $\hat{A}$ only contains edges that lead to child nodes, and hence $\hat{D}$ is computed based on this modified $\hat{A}$. TCPGen then takes the node encodings of the final layer, $H^\text{gcn}(L)$, to compute key and value vectors in the same way as tree-RNN. As a practical consideration for deep networks in general, residual connections and layer normalisation are added for any GNNs with multiple layers.

Although  lookahead with a configurable scope can be achieved by varying the number of layers $L$, recent research found that the performance of GCN starts to degrade with more than three layers.  
Apart from the fact that deeper networks are more difficult to train, it was pointed out that the representations of the nodes in GCN are inclined to converge to a certain value and hence become indistinguishable, which is referred to as the \textit{over-smoothing} problem. One promising method to address this problem is to build a shortcut directly linking to the first GCN layer to ensure a certain fraction of the final representation comes from the current node itself. Thus, GCNII is also investigated in this paper with each layer computed in Eqn \eqref{eq:gcnii}:
\begin{equation}
    H^\text{gii}(l+1) = f\Big{(}[(1-\alpha)\hat{P}H^\text{gii}(l)+\alpha H^\text{gii}(0)] W_\beta(l)\Big{)}
    \label{eq:gcnii}
\end{equation}
where $H^\text{gii}(l)$ is the $l$-th layer output of GCNII, 
hyper-parameter $\alpha$ scales the shortcut to the first layer, and $W_\beta(l)$ is the parameter matrix defined as:
\begin{equation}
    W_\beta(l) = (1 - \beta_l) I_N + \beta_l W(l)
    \label{eq:gcnii2}
\end{equation}
where $\beta_l$ is a layer-dependent hyper-parameter which is $\text{ln}(1/l+1)$ in this paper. 

Although GCNII has shown improved performance with deeper GCN of more than 4 layers, the maximum length in our biasing list is usually less than 10. With this depth for tree structures, the over-smoothing problem is less of a concern compared to the best structure of GCNII with 64 layers, and the network complexity is more problematic. Therefore, a simple parameter-sharing scheme is also proposed in this paper for deep GNNs to reduce network complexity. Specifically, the parameter matrices in the first $K$ layers are shared:
\begin{equation}
    W(1) = W(2) =...= W(K)
\end{equation}
In contrast to other complex graph structures, message-passing operations from child nodes to the root in a tree were similar across different layers. Therefore, having the same weight matrix representing this process effectively reduces the model complexity to a degree that is adequate for tree encoding. In particular, $K=L-1$ for GCN in this paper so that there are effectively two layer parameters to be trained while maintaining the depth of the network. The first $K$ layers act as universal message passing and the last layer performs a final information aggregation from neighbouring nodes.

\subsection{GraphSAGE with Max Pooling}

Node encodings for Tree-RNN and GCN are based on summation, whereas previous research \cite{maxpoolwe,maxpoolsubword} has found that using max pooling also achieves competitive performance for word representation based on subword units. Hence, as an alternative GNN structure, GraphSAGE with a max pooling aggregator function is studied in this paper. GraphSAGE is a multi-layer GNN with each layer performing an information aggregation over a sampled set of child nodes followed by an update to the representation of the current node. Although one of the innovations in GraphSAGE is fixed-size sampling, as the training time biasing list is already a sampled subset from the full biasing list, the sampling of GraphSAGE is omitted in this paper. The computation of each layer is
\begin{align}
\nonumber    &\mathbf{h}_{\mathcal{N}_i}(l+1)=\max(\{\sigma(W_1(l)\mathbf{h}^\text{sage}_{n_k}+\mathbf{b}(l)),\forall n_k\in \mathcal{N}_{j}\}) \\
    &\mathbf{h}^\text{sage}_{n_j}(l+1) = \sigma(W_2(l)\text{Concat}(\mathbf{h}_{\mathcal{N}_j}(l+1); \mathbf{h}^\text{sage}_{n_j}(l)))
\end{align}
where $\max(\cdot)$ and $\text{Concat}(\cdot)$ denote the element-wise max pooling and concatenation operators, $\mathcal{N}_j$ is the set of child nodes of node $n_j$. Although slightly better than GCN, GraphSAGE also degrades when adding more layers. Therefore, it is proposed in this paper to apply parameter-sharing for GraphSAGE, where both $W_1=W_1(1)=W_1(2)=...=W_1(L)$ and $W_2=W_2(1)=W_2(2)=...=W_2(L)$ are separately shared across all layers respectively.

TCPGen with GNN encodings still achieves high efficiency in handling large biasing lists. In training, with a large biasing list of 1000 words, TCPGen with a tree-RNN was 3.5 times slower than the standard AED or N-T model, with a negligible increase in space complexity. Among the three GNNs, GCN achieved the highest efficiency for training as its computation can be parallelised most, whereas the recursive computation in tree-RNN and the max-pooling in GraphSAGE hinder their training speed respectively. As a result, GCN in training is 2.5 times slower than the standard AED model, while GraphSAGE is 3 times slower. Moreover, by generating GNN encodings offline before decoding once the biasing list is available, the time and space complexity during inference is close to the standard AED or N-T for biasing lists of thousands of words.

\subsection{Combination of GNN Encodings}
\label{sec:multi}

Combinations \cite{rnntjoint,combination} of GCN and GraphSAGE are explored in this paper for tree encodings, as they are conceptually complementary. GCN adopts a spectral approach where the graph Laplacian is used to aggregate information, while GraphSAGE directly exploits the graph structure and performs a max-pooling aggregation. To exploit the complementarity between the two GNNs, both additive and multiplicative combination methods are investigated here. 

Additive combination performs a weighted sum of node encodings from GCN and GraphSAGE as the final node encodings before being processed by TCPGen (see Eqn. \eqref{comb1}).
\begin{equation}
    \mathbf{h}_{n_j}^\text{comb} = \alpha^\text{gcn}U_1\mathbf{h}_{n_j}^\text{gcn}+\alpha^\text{sage}U_2\mathbf{h}_{n_j}^\text{sage}.
    \label{comb1}
\end{equation}
$U_1$ and $U_2$ are two parameter matrices to rearrange the orders of the element in each GNN encoding as they may not encode information in the same order \cite{combination}. Note that $\alpha^\text{gcn} + \alpha^\text{sage} = 1$ are weights that are either fixed or predicted via attention. The attention calculation is performed on each node $n$ separately, as shown in Eqn. \eqref{combattn}.
\begin{equation}
    [\alpha^\text{gcn}_{i,n}, \alpha^\text{sage}_{i,n}] = \text{Softmax}(\mathbf{q}_i^T[\mathbf{h}_{n_j}^\text{gcn}, \mathbf{h}_{n_j}^\text{sage}])
    \label{combattn}
\end{equation}
where $\mathbf{q}_i$ is the same query vector used to calculate the TCPGen distribution. In this way, different sets of weights are assigned to different nodes at different decoder steps. 

The multiplicative combination is performed via a low-rank approximation of the bilinear pooling method. The combination is shown in Eqn. \eqref{comb2}
\begin{equation}
    \hat{\mathbf{h}}_{n_j}^\text{comb} = U_3(\tanh{(U_1\mathbf{h}_{n_j}^\text{gcn}}) \odot \tanh{(U_2\mathbf{h}_{n_j}^\text{sage})})
    \label{comb2}
\end{equation}
where $U_1$, $U_2$ and $U_3$ are parameter matrices, and $\odot$ is the element-wise product between two vectors. Following \cite{combination}, a shortcut connection from each individual GNN encoding was provided to form the final combined encoding for TCPGen, as shown in Eqn. \eqref{combshort}.
\begin{equation}
    \mathbf{h}_{n_j}^\text{comb} = \hat{\mathbf{h}}_{n_j}^\text{comb} + U_4\mathbf{h}_{n_j}^\text{gcn} + U_5\mathbf{h}_{n_j}^\text{gcn}
    \label{combshort}
\end{equation}
where $U_4$ and $U_5$ are another two parameter matrices. 

\section{Experimental setup}
\label{sec:exp}

\subsection{Data}

Experiments were conducted on two distinct datasets, namely the LibriSpeech audiobook corpus and the AMI meeting data, where the latter followed the audio-visual speech recognition pipeline. The LibriSpeech corpus \cite{LibriSpeech}, consists of 960 hours of read English from audiobooks, was used for evaluation purposes, with the dev-clean and dev-other sets used for validation, while test-clean and test-other were employed for evaluation. To investigate the impact of critical hyper-parameters, small-scale experiments were carried out using the train-clean-100 subset as the training set. Moreover, models trained on the LibriSpeech dataset were fine-tuned and evaluated on the AMI dataset in accordance with the approach proposed in \cite{tcpgengnn}.

The AMI meeting corpus \cite{ami} consists of 100 hours of meeting recordings involving 4-5 individuals, which were divided into the train, dev, and eval sets. To demonstrate the effectiveness of contextual biasing on data from another domain with limited training resources, a subset comprising 10\% of the utterances from the AMI training set corresponding to 8 hours of audio was used to fine-tune the models previously trained on the LibriSpeech 960-hour data. There were 14 meetings from the dev set and 8 meetings from the eval set accompanied by slides that were selected to formulate the new test set for the audio-visual contextual ASR pipeline.

The 80-dim FBANK features at a 10~ms frame rate concatenated with 3-dim pitch features were used as the model input. SpecAugment \cite{specaug} with the setting $(W,F,m_F,T,p,m_T)=(40,27,2,40,1.0,2)$ was used without any other data augmentation or speaker adaptation.

\begin{figure*}[t]
    \centering
    \includegraphics[scale=0.24]{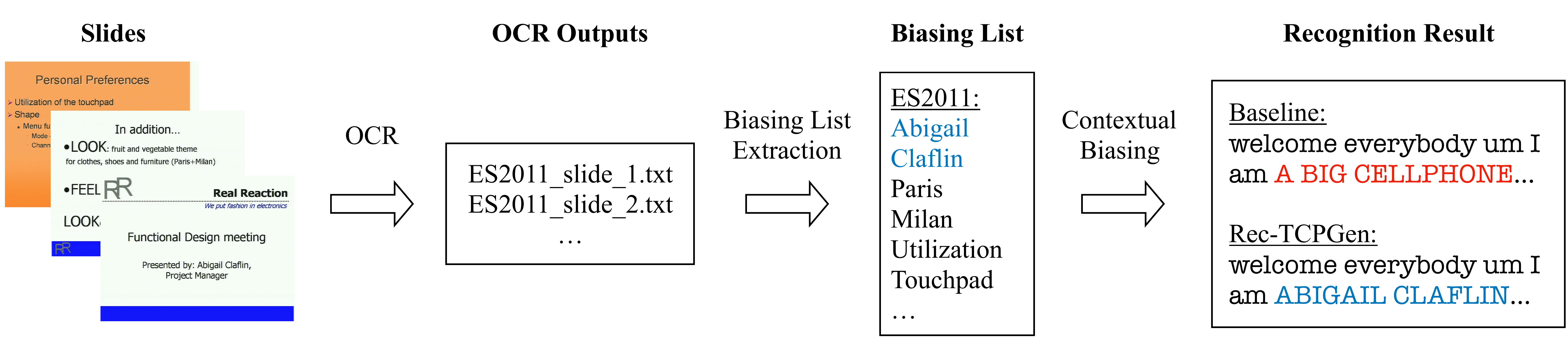}
    \vspace{-0.3cm}
    \caption{Illustration of the visual-grounded contextual ASR pipeline for the meeting series ES2011 containing meetings ES2011a to ES2011d.}
    \label{fig:amisetup}
    \vspace{-0.3cm}
\end{figure*}

\subsection{Biasing list selection}
To simulate real-world scenarios in LibriSpeech, the complete list of rare words comprising 200,000 distinct words as suggested in \cite{DBRNNT} was used. The full rare word list consisted of over 60\% out-of-vocabulary (OOV) words that were absent from the LibriSpeech speech training set. Consistent with the method proposed in \cite{DBRNNT}, the biasing lists were extracted by identifying words from the full rare word list that appeared in the reference transcription of each utterance, followed by the addition of a specific number of distractors. During inference, 10.3\% of the word tokens in the test sets belonged to the full rare word list.

Fig. \ref{fig:amisetup} shows the visual-grounded contextual ASR pipeline for AMI that utilises optical character recognition (OCR) output for slides. The Tesseract 4 OCR engine, equipped with LSTM models\footnote{OCR implementation at \url{https://github.com/tesseract-ocr/tesseract}}, was first applied to the slides of each meeting series (e.g., ES2011[a-d]). Subsequently, distinct word tokens were extracted from the OCR output text files, and words in the full rare word list, which also occurred fewer than 100 times in the AMI training set, were selected to form the biasing list for that particular meeting series. These meeting-specific biasing lists were then used for the recognition of all utterances in that meeting series. The sizes of the biasing lists vary between 175 to 576, and the total number of word tokens covered by these lists was 1,751 out of 112,110 word tokens (1.5\%). As shown in Fig. \ref{fig:amisetup}, these words mainly consisted of highly valuable content words whose accurate recognition was crucial for comprehending the utterance. Therefore, although the biasing lists had a minor impact on the overall word error rate (WER), they were essential for improving the recognition performance of critical words.
Details of the meetings with slides and the extraction pipeline are available\footnote{\url{https://github.com/the-anonymous-bs/AMIslides\_biasing}}.

\subsection{Model specification}
The ESPnet toolkit \cite{espnet} was used for developing the systems. A unigram word piece model comprising 600 unique word pieces was created on the LibriSpeech data and was applied directly to the AMI data. Both the AED and N-T models employed a Conformer \cite{conformer} encoder, which comprised 16 conformer blocks comprising 4 attention heads of size 512. The AED used a single-layer LSTM decoder of size 1024 and a location-sensitive attention mechanism featuring 4 heads of size 1024. The N-T, on the other hand, employed a 1024-dimensional predictor and a joint network consisting of a single fully-connected layer of size 1024. GNN encodings for LibriSpeech train-clean-100 experiments used 256-d GNN encodings, whereas the LibriSpeech full-scale experiments used 1024-d for GNN encoders.

LM shallow fusion and BLMD were implemented using a two-layer LSTM-LM with 2048 hidden units trained on the 800 million-word text training corpus of LibriSpeech as the target domain LM for the LibriSpeech experiments. Each source domain LM, trained on the text of the audio training data, used a single-layer LSTM with 1024 hidden units. It is worth noting that each LM had the same word pieces as the corresponding ASR system.

\subsection{Training specifications}
During training, biasing lists with 1000 distractors were used for the experiments conducted on the LibriSpeech dataset, while 100 distractors were used for the AMI data. To create these lists, biasing words were selected from the reference transcription, and additional distractors were added. To prevent the AED model from becoming overly confident about TCPGen outputs, a dropout-inspired technique was employed during training, as described in \cite{DBRNNT}. Specifically, biasing words that were presented in the reference transcription had a 30\% probability of being removed from the biasing list. The Conformer was optimised using the Noam optimiser \cite{transformer}. Additionally, the hyper-parameters for the BLMD model were determined based on the respective dev sets for each dataset.

\vspace{-0.05cm}
\subsection{Evaluation metrics}
\vspace{-0.05cm}
In addition to WER, the rare word error rate (R-WER) was used to evaluate the system performance on biasing words that were ``rare" in the training data for that system. R-WER is the total number of \textit{error} word tokens that belong to the biasing list divided by the total number of word tokens in the test set that belong to the biasing list. Insertion errors were counted in R-WER if the inserted word belonged to the biasing list, in contrast to \cite{MEM}. In addition, OOV WER was also computed in the same way as R-WER but for OOV words in the biasing list. There are altogether 443 such words in LibriSpeech test-clean and test-other sets. Moreover, the slides' rare word error rate (R$_\text{s}$-WER) is reported for the AMI experiments calculated in the same way as R-WER, but for the rare words in slides. Insertions of slides biasing words were included in R$_\text{s}$-WER. 

As rare words are scarce in the dataset, significance tests were performed to ensure that the improvements found by using GNN encodings were statistically significant. Specifically, independence was assumed at the book level for LibriSpeech and at the speaker level for AMI. The alternative hypothesis was defined as the GNN system performing better than the standard TCPGen (i.e. one-tailed sign test). 

\section{Results}
\label{sec:results}

\subsection{LibriSpeech train-clean-100 Results}

\begin{figure*}[t]
    \centering
    \includegraphics[scale=0.46]{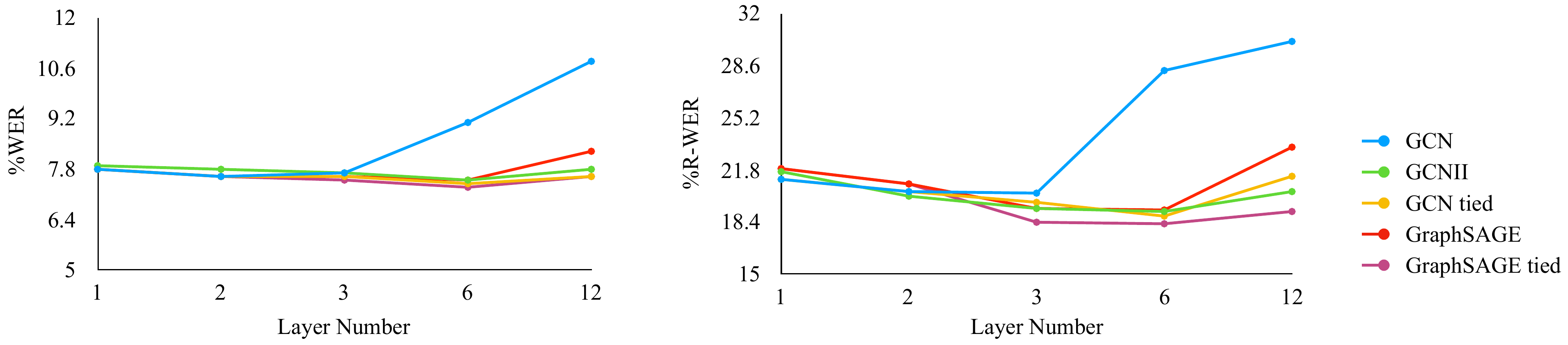}
    \vspace{-0.3cm}
    \caption{Plot of WER (\%) and R-WER (\%) against the number of GNN layers for N-T on LibriSpeech test-clean data. Systems were trained on train-clean-100 for 120 epochs. Biasing lists with 1000 distractors were used. ``Tied" referred to the parameter-tying scheme.}
    \label{fig:rnntlayer}
\end{figure*}

\begin{figure*}[t]
    \centering
    \includegraphics[scale=0.46]{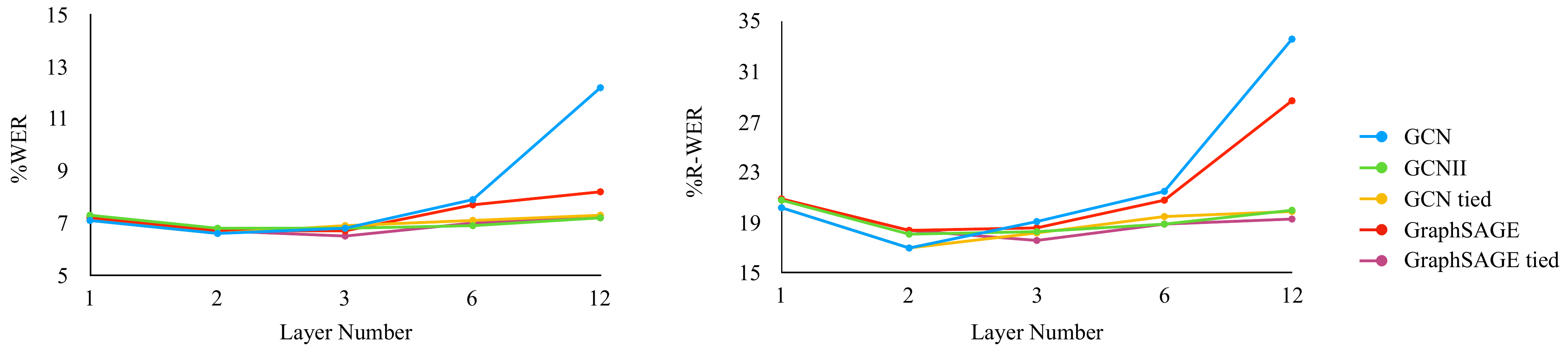}
    \vspace{-0.3cm}
    \caption{Plot of WER (\%) and R-WER (\%) against the number of GNN layers for AED on LibriSpeech test-clean data. Systems were trained on train-clean-100 for 120 epochs. Biasing lists with 1000 distractors were used. ``Tied" referred to the parameter-tying scheme.}
    \label{fig:aedlayer}
\end{figure*}

\begin{figure*}[t]
    \centering
    \includegraphics[scale=0.5]{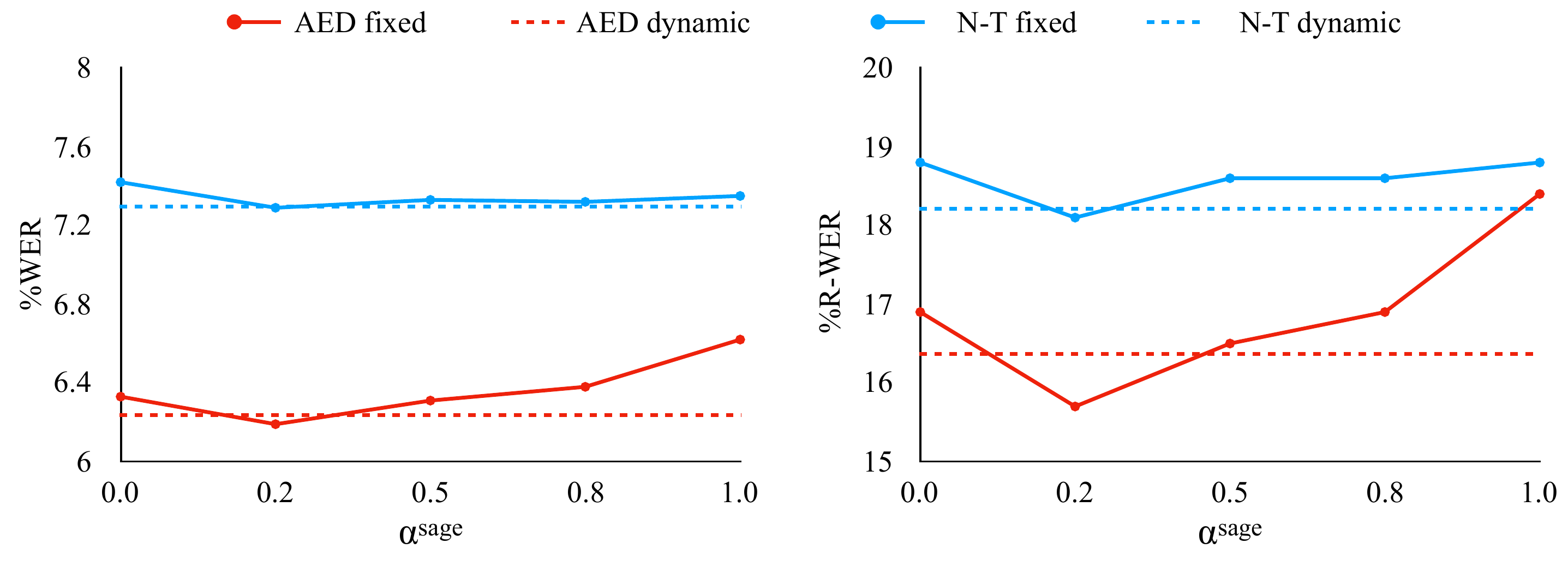}
    \vspace{-0.5cm}
    \caption{Variation of WER and R-WER against model combination weight $\alpha_\text{sage}$, and dynamic refers to using attention mechanism for weight calculation.}
    \label{fig:combweights}
\end{figure*}

Experiments were first performed on the train-clean-100 set to find the best-performing GNN setups. The first investigation was on the number of GNN layers which determines how much lookahead is needed. The plots of WER and R-WER against the number of layers for N-T are shown in Fig. \ref{fig:rnntlayer}, and those for AED are shown in Fig. \ref{fig:aedlayer}. Note that parameter tying only had effects when the layer number exceeded two.

For N-T, both GCN and GraphSAGE had a clear trend that the performance tended to degrade when the number of layers was increased beyond three and degraded significantly when it reached twelve layers. GraphSAGE inherently suffered less from this problem than GCN as the max pooling operator enabled the gradient for salient nodes to remain at its original value instead of being over-smoothed \cite{gcnii}. This degradation was mitigated by either using the GCNII structure or the parameter-tying scheme proposed in this paper. The best performance was achieved by 6-layer systems using WER as the selection criterion, where GCN and GraphSAGE with tied parameters achieved better performance than GCNII. 

Similar observations were found with AED, where GCN and GraphSAGE with tied parameters achieved slightly better performance than GCNII. However, the best performance was achieved using 2-layer GCN and 3-layer GraphSAGE models. This difference is mainly caused by the label-synchronous nature of AED, where the model required knowledge for predicting only the next token, and information further into the future is less useful for this prediction in contrast to N-T.

The best-performing GCN and GraphSAGE with tied parameters were used for model combination. For additive combination, different fixed combination weights gave results shown in Fig. \ref{fig:combweights}, together with dynamic combination weights. As a result, dynamic combination performed better than most fixed-weight combinations, whereas the best performance was still obtained by the fixed-weight combination for both AED and N-T, with $\alpha_\text{sage}=0.2$. By examining the predicted dynamic weights, it was found that unless at the root node of the tree, the dynamic weight almost completely ignored GraphSAGE encodings and hence suffered from mode collapse, and hence GraphSAGE encodings were not properly trained. In fact, since GCN is usually better at handling near-future information, the model learnt to only rely on GCN.


\subsection{LibriSpeech 960-hour Results} 
\begin{table*}[t]
    \centering
    \caption{WER and R-WER on LibriSpeech test-clean and test-other sets using Conformer \textbf{AED} and TCPGen trained on LibriSpeech 960-hour data with various GNN encodings. Note that both GCN (2-layer) and GraphSAGE (3-layer) adopted parameter tying. GCN and additive combination (fixed) were selected as the representative GNNs to be evaluated with BLMD.}
    \begin{tabular}{llccccccc}
    \toprule
    & & & \multicolumn{3}{c}{test-clean (\%)} & \multicolumn{3}{c}{test-other (\%)} \\
    System     & GNN Enc. & BLMD & WER & R-WER & OOV WER & WER & R-WER & OOV WER \\
    \midrule
    Conformer AED & N/A & $\times$ & 3.71 & 13.2 & 71.2 & 9.36 & 29.5 & 75.5 \\
    ~~~~+TCPGen & No & $\times$ & 3.34 & 8.4 & 40.1 & 8.43 & 21.3 & 41.5 \\
    ~~~~+TCPGen & Tree-RNN & $\times$ & 3.13 & 6.7 & 33.8 & 7.94 & 17.8 & 34.7 \\
    ~~~~+TCPGen & GCN tied & $\times$ & 2.81 & 6.7 & 32.9 & 7.34 & 17.1 & 33.6 \\
    ~~~~+TCPGen & GCNII &$\times$ & 2.88 & 6.8 & 34.1 & 7.46 & 17.5 & 34.7 \\
    ~~~~+TCPGen & GraphSAGE tied & $\times$ & 2.91 & 6.6 & 32.1 & 7.42 & 17.0 & 33.1 \\
    ~~~~+TCPGen & Additive Combination (fixed) & $\times$ & {2.59} & {5.8} & 31.8 & 7.00 & 15.5 & 30.5 \\
    ~~~~+TCPGen & Additive Combination (dynamic) & $\times$ & 2.72 & 6.3 & {31.5} & 7.17 & 16.4 & 31.8 \\ 
    ~~~~+TCPGen & Bilinear Combination & $\times$ & 2.62 & 6.1 & 32.3 & 7.08 & 16.1 & 31.4\\
    \midrule
    Conformer AED & N/A & $\checkmark$ & 3.33 & 12.3 & 69.3 & 8.04 & 27.6 & 74.2 \\
    ~~~~+TCPGen & No & $\checkmark$ & 2.79 & 6.9 & 31.3 & 7.40 & 19.5 & 32.5 \\
    ~~~~+TCPGen & GCN tied & $\checkmark$ & 2.48 & 5.2 & 28.1 & 6.33 & 14.2 & 27.7 \\
    ~~~~+TCPGen & Additive Combination (fixed) & $\checkmark$ & \textbf{2.26} & \textbf{4.2} & \textbf{23.9} & \textbf{5.87} & \textbf{11.5} & \textbf{23.0}\\
    \bottomrule
    \end{tabular}
    \label{tab:aedfull}
\end{table*}

\begin{table*}[t]
    \centering
    \caption{WER and R-WER on LibriSpeech test-clean and test-other sets using Conformer \textbf{N-T} and TCPGen trained on LibriSpeech 960-hour data with various GNN encodings. Note that both GCN (6-layer) and GraphSAGE (6-layer) adopted parameter tying. GCN and bilinear combination were selected as the representative GNNs to be evaluated with BLMD}
    \begin{tabular}{llccccccc}
    \toprule
    & & & \multicolumn{3}{c}{test-clean (\%)} & \multicolumn{3}{c}{test-other (\%)} \\
    System & GNN Enc. & BLMD & WER & R-WER & OOV WER & WER & R-WER & OOV WER\\
    \midrule
    Conformer N-T & N/A & $\times$ & 4.02 & 14.1 & 80.1 & 10.12 & 33.1 & 83.2 \\
    ~~~~+TCPGen & No & $\times$ & 3.40 & 8.9 & 43.3 & 8.79 & 22.2 & 46.7 \\
    ~~~~+TCPGen & Tree-RNN & $\times$ & 3.14 & 7.6 & 40.6 & 8.23 & 18.8 & 45.3 \\
    ~~~~+TCPGen & GCN tied & $\times$ & 3.11 & 7.0 & 39.1 & 8.14 & 18.4 & 43.7 \\
    ~~~~+TCPGen & GCNII & $\times$ & 3.16 & 7.7 & 40.1 & 8.36 & 18.9 & 45.2 \\
    ~~~~+TCPGen & GraphSAGE tied  & $\times$& 3.10 & 6.9 & 39.1 & 8.18 & 18.6 & 44.2 \\
    ~~~~+TCPGen & Additive Combination (fixed) & $\times$ & 2.99 & 6.5 & 37.5 & 8.10 & 16.7 & 38.9 \\
    ~~~~+TCPGen & Additive Combination (dynamic) & $\times$ & 3.02 & 6.6 & 39.1 & 8.14 & 17.2 & 40.3 \\
    ~~~~+TCPGen & Bilinear Combination & $\times$ & 2.97 & 6.2 & 36.9 & 8.02 & 16.6 & 38.1 \\
    \midrule
    Conformer N-T & N/A & $\checkmark$ & 3.55 & 12.5 & 78.2 & 8.90 & 30.4 & 81.8 \\
    ~~~~+TCPGen & No & $\checkmark$ & 3.02 & 8.0 & 40.6 & 7.49 & 18.6 & 43.7 \\
    ~~~~+TCPGen & GraphSAGE tied & $\checkmark$ & 2.71 & 6.3 & 38.0 & 7.34 & 17.7 & 44.9 \\
    ~~~~+TCPGen & Bilinear Combination & $\checkmark$ & \textbf{2.56} & \textbf{5.3} & \textbf{34.9} & \textbf{6.89} & \textbf{14.1} & \textbf{34.8} \\
    \bottomrule
    \end{tabular}
    \label{tab:rnntfull}
\end{table*}

The main results for LibriSpeech full-set experiments were summarised in Table \ref{tab:aedfull} for AED and in Table \ref{tab:rnntfull} for N-T respectively. Compared to the standard TCPGen, all three types of GNNs achieved significantly better WER and R-WER (at p values smaller than 0.01) on both test sets for both AED and N-T. In particular, using multi-layer GNN, such as GCN and GraphSAGE, achieved clearly better performance to tree-RNN on AED, whereas the performance difference among those three GNN types was less obvious on N-T. The best-performing GNN structure on both AED and N-T was GCN with tied parameters. For AED, GCN achieved 16\% relative WER reduction with a 20\% R-WER reduction on the test-clean set and 13\% relative WER reduction with 19\% relative R-WER reduction on the test-other set (comparing row 4 to row 2 in Table \ref{tab:aedfull}). For N-T, GCN achieved a 9\% relative WER reduction with 21\% relative R-WER reduction on the test-clean set, and a 7\% WER reduction with a 17\% relative R-WER reduction on the test-other set (comparing row 4 to row 2 in Table \ref{tab:rnntfull}). With similar levels of R-WER reduction, AED achieved a higher reduction in WER. As analysed in \cite{tcpgenmbr}, TCPGen produced a much more confident prediction of $P^\text{gen}$ with AED than N-T, where the main reductions in overall WER were attributed to the reduction in R-WER. The improvements using GNN indicated that the GNN encoding improved the prediction of $P^\text{gen}$, which was more beneficial for the overall WER in AED.

Both additive and bilinear combinations of GNN encodings achieved superior performance to individual GNN encodings with both AED and N-T models. For the AED model, the best performance was achieved by the additive combination with the best set of fixed weights previously found on train-clean-100 experiments, while the bilinear combination achieved very similar performance to the additive one. This led to a total of 31\% relative R-WER reduction compared to the standard TCPGen (comparing row 7 to row 2 in Table \ref{tab:aedfull}), and a total of {56\%} relative R-WER reduction compared to the baseline Conformer AED model. The performance improvement with the GNN combination was smaller on N-T, with the best results achieved by the bilinear pooling combination. This resulted in a 30\% relative R-WER reduction compared to the standard TCPGen (comparing row 9 to row 2 in Table \ref{tab:rnntfull}), and an overall {56\%} relative R-WER reduction compared to the baseline Conformer N-T model. 

Finally, selected systems were evaluated with BLMD, where TCPGen with GNN encodings achieved further performance improvements. The best-performing system for AED was {TCPGen} with the additive combination of GNN encodings, which achieved an overall 66\% relative R-WER reduction on the test-clean set and 58\% reduction on the test-other compared to the baseline. For N-T, an overall 57\% relative R-WER reduction was achieved on test-clean, and 54\% was achieved on test-other.
Moreover, the OOV-WER had the same reduction pattern as R-WER for both AED and N-T. The best AED and N-T systems with combined GNN encodings for TCPGen reduced the OOV-WER by over 60\%. Notably, BLMD was particularly beneficial for GNN encodings in AED systems, reducing the OOV-WER to {1/3} of the baseline value. This confirmed that even though GNN required more parameters to encode the prefix-tree, TCPGen still generalises well to unseen branches (i.e. OOV words) on the tree.

\subsection{AMI Audio-visual Contextaul ASR experiments}
\begin{table}[h]
    \centering
    \caption{WER (R$_s$-WER) on AMI test set using the audio-visual contextual ASR pipeline with 10\% of AMI training set. Baseline is the standard AED and N-T systems, and AMI baseline (the first row) specifically refers to the standard system trained from scratch on the full AMI training set.}
    \begin{tabular}{llccccc}
    \toprule
    System & GNN Enc. & BLMD & AED (\%) & N-T (\%) \\
    \midrule
    AMI Baseline & N/A & $\times$ & 23.6 (56.3) & 26.5 (58.0) \\
    \midrule
    Baseline & N/A & $\times$ & 22.2 (51.2) & 25.7 (52.6) \\
    ~~~~+TCPGen & No & $\times$ & 22.0 (40.5) & 25.5 (44.7) \\
    ~~~~+TCPGen & Tree-RNN & $\times$ & 21.9 (36.7) & 25.4 (40.7) \\
    ~~~~+TCPGen & GCN tied & $\times$ & 21.9 (35.3) & 25.4 (40.7) \\
    ~~~~+TCPGen & GraphSAGE tied  & $\times$& 21.9 (36.4) & 25.2 (39.6)\\
    ~~~~+TCPGen & Additive Comb. & $\times$ & 21.8 (33.1) & 25.2 (37.2) \\
    ~~~~+TCPGen & Bilinear Comb. & $\times$ & 21.8 (33.5) & 25.1 (36.2) \\
    \midrule
    Baseline & N/A & $\checkmark$ & 21.1 (45.5) & 24.3 (46.5) \\
    ~~~~+TCPGen & No & $\checkmark$ &  20.9 (34.2) & 24.1 (37.7) \\
    ~~~~+TCPGen & GCN tied & $\checkmark$ & 20.8 (32.2) & 23.7 (35.8)  \\
    ~~~~+TCPGen & Best Comb. & $\checkmark$ & \textbf{20.7} (\textbf{29.7}) & \textbf{23.6} (\textbf{32.8}) \\
    \bottomrule
    \end{tabular}
    \label{tab:amifull}
\end{table}

The performance of various GNN encodings for TCPGen was further integrated into the audio-visual contextual ASR pipeline, and results were shown in Table \ref{tab:amifull}. In general, reductions in R-WER had a much smaller influence on the overall WER than with LibriSpeech, as rare words only occupied a much smaller portion. The findings were consistent with LibriSpeech, where the best-performing combination methods for AED and N-T were additive and bilinear respectively. However, the relative R-WER improvement was smaller compared to that in the LibriSpeech experiments, as the best-performing system here already had an R-WER that was very close to the overall WER whereas the R-WER in LibriSpeech was still twice as high as the overall WER. Compared to the baseline standard systems, TCPGen in AED achieved over 35\% relative R-WER reduction using the best combined GNN encodings, while TCPGen with the best combined GNN encodings in N-T achieved over 30\% relative R-WER reduction both with and without BLMD.

\section{Conclusion}
\label{sec:conclusion}
This paper proposes three different types of GNN encodings in TCPGen for end-to-end contextual ASR, including tree-RNN, GCN and GraphSAGE. GNN encodings gave a lookahead functionality by incorporating future information on branches starting from the current node. Combination methods that take advantage of the complementarity between GCN and GraphSAGE were also explored. Experiments on LibriSpeech and AMI following an audio-visual contextual ASR pipeline showed consistent and significant WER and R-WER improvement for both AED and N-T systems using GNN encodings. The best combined GNN encodings achieved over {60}\% R-WER and OOV-WER reductions compared to the baseline standard systems.



\vfill

\end{document}